# ENTROPY-DIFFERENCE BASED STEREO ERROR DETECTION

*Subhayan Mukherjee\*, Irene Cheng\*, Ram Mohana Reddy Guddeti[+]* and *Anup Basu\**

\*University of Alberta, Edmonton, Canada   [+]National Institute of Technology Karnataka, India

**ABSTRACT**

Stereo depth estimation is error-prone; hence, effective error detection methods are desirable. Most such existing methods depend on characteristics of the stereo matching cost curve, making them unduly dependent on functional details of the matching algorithm. As a remedy, we propose a novel error detection approach based solely on the input image and its depth map. Our assumption is that, entropy of any point on an image will be significantly higher than the entropy of its corresponding point on the image's depth map. In this paper, we propose a confidence measure, Entropy-Difference (ED) for stereo depth estimates and a binary classification method to identify incorrect depths. Experiments on the Middlebury dataset show the effectiveness of our method. Our proposed stereo confidence measure outperforms 17 existing measures in all aspects except occlusion detection. Established metrics such as precision, accuracy, recall, and area-under-curve are used to demonstrate the effectiveness of our method.

## 1. INTRODUCTION

Stereo disparity estimation is an important, well-researched topic, and has applications in numerous areas, viz. robotic vision, 3D scene reconstruction, object detection and tracking [1]. Its main challenge is generating accurate depth information of a scene by comparing the left and right image pixels, with only color and spatial information, i.e., the low level image features [2]. Stereo depth estimation has progressed substantially in the past two decades. So, now, many effective and efficient stereo disparity estimation methods exist, which are classified in [3]. In particular, dense stereo matching can improve the bottom-up stereo processing chain significantly [4]. Dense algorithms often compute an initial (noisy) depth map of the scene first, using some baseline methods; they then use regularizing methods [5] to effectively smoothen areas of little depth variance. Further, w.r.t. detecting high-quality yet sparse seed depth estimates and then computing dense depth maps, authors' contributions are often placed jointly on detection and completion [6]. However, for many strong stereo algorithms, unreliable estimates are more due to physical reasons (e.g., occlusions) rather than matching incapability for the visible parts in two views (e.g., faced by simple box filter methods). Thus, many advanced methods model and handle occlusions, bypassing the classification of initial depth estimates of low quality [7].

Stereo depth estimation, based on successive refinement of (initial) depth estimates, like [5, 8] represents an entire class of depth refinement approaches which classify a pixel-wise depth map into two categories, reliable and unreliable, followed by refining of pixels with unreliable depth values. For this, often, left-right consistency check (LRC), a popular and handy choice for detecting inconsistent depth estimates, is used. Depth map refinement techniques [9] are classified as pre-filtering and reliability-based approaches. The former uses Adaptive filters, Asymmetric Gaussian filters, etc. to smoothen depth maps; the latter uses reliable warping information from multiple views for depth map hole-filling.

We propose a novel classification method for such initial disparity estimates, and compare it with LRC. In the corresponding experiments, we use a simple, generic, block-based stereo matching method based on the Sum of Absolute Differences (SAD) cost function and following the Winner-Takes-All (WTA) strategy to obtain the initial disparity estimates mentioned earlier. We further experiment on depth maps produced by very recent, advanced stereo algorithms. Additionally, we propose a novel stereo confidence measure and compare it with 17 existing ones, following [10].
However, a certain aspect of our confidence measure makes it more challenging compared to ones outlined in [10], and this should be kept in mind while interpreting all our experimental results. All the surveyed measures base their computations on characteristics of the "matching cost curve," and thus, use a lot of additional information as inputs, which are not required by our measure. We estimate disparity errors only by comparing the supplied depth map, and the original image, without utilizing any "extraneous" inputs, (e.g., "matching cost curves") about the (preceding) stereo matching process.

It should also be carefully noted that the Negative Entropy Measure (NEM) [10] is not similar to our confidence measure in any way. In NEM, pixel matching cost values are converted to a probability density function (pdf), whose negative entropy is used as a measure of confidence. On the other hand, in our measure, difference of entropies between image pixels and depth map pixels is used to measure the confidence. Similarly, Number of Inflection Points (NOI) [10] is different from our measure, since it measures the number of minimum valleys in matching cost curves, whereas *our measure* does not involve any inflection points at all. Our proposed threshold detection technique uses the single inflection point of a 3rd degree polynomial approximating (not the matching cost curve, but) the "entropy variations in Ent_D vs. percentiles" curve (which is also specific to our method).

Finally, the confidence measures studied by the authors in [10] estimate the confidence of disparity assignments by looking at the matching cost curves of only the individual pixels, without considering their neighboring pixels or any other sort of global information. On the contrary, our measure calculates the entropies of square blocks of pixels centered on every image pixel. Thus, the values of the neighboring pixels of any given pixel are also considered, producing more reliable results.

In recent years, some supervised-learning based stereo error detection approaches have been developed using ensemble learning of multiple confidence measures [11] or learning several extracted features for each pixel, including the cost curves for a pixel and its neighbors [12]. Hence, they are inherently dependent on much more information beyond the raw image and its depth map, and are thus not comparable with respect to input modality or performance.

## 2. PROPOSED METHOD

Fig. 1 shows the flowchart of our approach w.r.t computing the confidence measure. This measure is subsequently used as input to our "Threshold Detection" phase, described later.

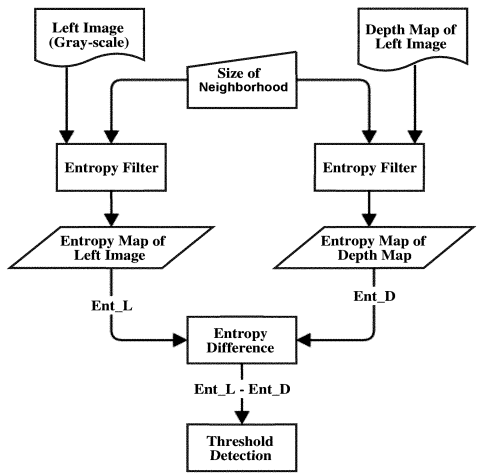

Fig. 1. Flowchart of our Proposed Method. Pixels of Ent_L and Ent_D correspond to entropies of left image and depth map pixels respectively. Ent = Ent_L – Ent_D

Our confidence measure computes the difference in local entropies of each pixel of left (or right) image of the stereo pair, and its depth map, both of which are taken as inputs.

### A. Inputs
The inputs to our proposed algorithm are the following:
1. Left Image (Gray-scale): Left image of the input stereo pair is converted to 'Lab' color space. Only the lightness (L) value is processed, and the chrominance (a, b) are discarded.
2. Depth Map of Left Image: The depth map generated by the dense stereo correspondence algorithm, w.r.t left image of the stereo pair is the 2nd input to our algorithm. This depth map has incorrect disparity estimates for some of the image pixels; our thresholding algorithm attempts to identify them.

It should be noted that though we use the left image and its depth map in our experiments, the proposed work has not been developed specifically for the left image, rather it is equally applicable to the right image and its depth map.
3. Size of Neighborhood: This is required for our entropy computation step, where Shannon's entropy (separately) is calculated for each pixel of left image, and its depth map, by considering a square area around a given pixel. Said pixel always occupies the central position in this square area; so, the value supplied for 'size of neighborhood' must be odd.

### B. Entropy Computation
Entropy is a statistical measure of randomness that can be used to characterize the texture of the input image. We use symmetric border padding, so all values of padding pixels are mirror reflections of border. Shannon's entropy measure, (Eq. 1) is computed for pixels of input image and depth map.

$$h = -\sum_{i=1}^{n} p_i . * \log_2 p_i \qquad (1)$$

Here, $p_i$ contains the normalized histogram counts of the neighborhood around any given pixel. The number of histogram bins is set to 256 (gray-scale image). We compute such entropies for both the (gray-scale) left image, and the depth map output by the stereo correspondence algorithm. Thus, we obtain two entropy maps. One is Ent_L (for the left image), and the other for the depth map, denoted by Ent_D.

### C. Core assumption for stereo error detection
Our method's *core* assumption is that, *entropy of any point on an image will be significantly higher than the entropy of its corresponding point on the depth map of that image*.
This assumption is based on the observation that 'L' values (lightness / illumination) can vary significantly (compared to the depth level) in any pixel's neighborhood due to multiple reasons: image noise, differences in lighting (e.g., shadows or reflections), differences in surface texture etc. even when the neighboring pixels lie on the same surface and thus, roughly at the same depth level. Based on this, we now subtract the value of each pixel of the depth map's entropy map, Ent_D, from the corresponding pixel of the left image's entropy map, Ent_L, to generate the entropy difference image, 'Ent' as the output. Some of Ent's pixels may have negative values (we do not compute an absolute difference of entropy). So, high and positive 'Ent' values correspond to correct depths.

**D. Threshold Detection**

This step deals with deriving a pixel value threshold, Ent_Th, from the entropy difference image, 'Ent'. All pixels in the depth map of the left image corresponding to those in 'Ent' with values less than Ent_Th will be classified as incorrect disparity estimates. The detailed steps are:

**Algorithm 1** Proposed Threshold Detection Method
| | |
|---|---|
| 1: | Compute percentiles $P_i$ over **Ent**, $\forall$ **i** = 1, 2 … 100 |
| 2: | $\forall$ $P_i$, compute $E_i = \sigma$ {ent_$d_{xy}$ \| ent$_{xy}$ < $P_i$} |
| 3: | Compute a $3^{rd}$ degree polynomial fit, $f$, estimating $E_i$ given $P_i$ |
| 4: | Determine the percentile, $f_p$ w.r.t inflection point of $f$ |
| 5: | If $f_p \in [P_{20}, P_{80}]$, Then |
| 6: |     Set **Ent_Th** := $f_p$ |
| 7: | Else |
| 8: |     Set **Ent_Th** := $P_{50}$ |
| 9: | End |

A key step in Algorithm 1 is step 2 where for each percentile $P_i$ over Ent, we compute the standard deviation of entropies of depth map pixels for which the corresponding input image entropies are lesser than the said percentile. It may be noted that we are using a $3^{rd}$ degree polynomial so as to obtain a single, unique inflection point. Also, in case $f_p$ does not lie between $P_{20}$ and $P_{80}$, we take Ent_Th to be $P_{50}$, so as to disallow extreme values of Ent_Th (entropy threshold). This is one particular scenario where our method fails to detect any "ideal" threshold. Our approach is based on the observation that regions of wrong depth estimates are often comprised of random, uncorrelated disparity values concentrated in small spatial regions; entropies of "unstable" regions of the depth map (like these) are expected to be high. Again, regions of low entropy difference ('Ent') values translate to spaces of incorrect disparity estimates. Thus, combining the above observations, we chose to examine, how (by gradually taking higher values of 'Ent' or $P_i$ as the threshold, Ent_Th) the entropy of depth estimates corresponding to 'Ent' values less than Ent_Th (or, $P_i$) vary (measured by standard deviation, '$\sigma$' of the entropy values), as shown above, in Algorithm 1.

## 3. RESULTS AND DISCUSSION
**A. Performance evaluation of confidence measure**
For a comprehensive evaluation of our proposed confidence measure, we compared with [10], where the authors were able to demonstrate the better performance of their method by comparing with the performance data of 17 established confidence measures presented in Tables 1, 3, 4 and 6 [10]. Although few recent approaches using ensemble learning of multiple confidence measures [11] and per-pixel feature learning / extraction [12] have been developed, they require much more information in addition to the raw image and its depth map, and are thus not comparable with our measure with respect to input modality or performance.
We use the quality metric 'Area-Under-Curve' (AUC) which measures the likelihood of a confidence measure to predict correct matches by varying neighborhood sizes. We closely follow the methodology outlined by the authors in [10]. Similar to them, we experimented on 29 Middlebury dataset stereo pairs published between 2002 and 2007. Because our proposed stereo confidence measure's operational model is entirely different compared to the ones existing in literature [10], so, not all experiments done in [10] are relevant for us:
1. Similar to the other authors, we follow a WTA strategy with the SAD cost function (taking square windows of same sizes as the authors) to generate the stereo depth maps.
2. While reporting scores for our confidence measure, we average our calculations over all disparity maps and all SAD aggregation window sizes (which in our case is also same as neighborhood sizes) used by the authors following the exact same methodology. We use the same number of steps as the authors for approximating our AUC calculations.
3. As our confidence measure is not based on the matching cost curves, we do not conduct any additional experiments using NCC cost curves in place of their SAD counterparts.
4. Unlike the authors, we do not need to test the usefulness of proposed measure for selection of the true disparity among hypotheses generated by different matching strategies, as our measure does not take these hypotheses as input (since it is oblivious of any details regarding the stereo matching process used to generate the supplied depth maps).
5. To meet space constraints of this paper, we performed all our experiments on binocular stereo images in the rectified canonical configuration, but not on multi-baseline imagery or additional datasets on real-world stereo pairs, e.g. KITTI.
6. We were unable to perform relevant comparisons of our confidence measure w.r.t Table 2 and 5 of [10], as those tables mention only ranks of the surveyed methods, without revealing actual values of their corresponding performance metrics used by authors for the comparisons and rankings.
For Table 1 [10], the lowest AUC achieved by any method on Teddy is 0.075 by LRD, but our proposed measure achieved an even lower (i.e., better) AUC of 0.066, for the same window size of 11 × 11 pixels. In Table 3 [10], best performance value is obtained using LRD, but our proposed measure achieved 0.478, thus surpassing all others. In Table 4 [10], again, LRD is the top performer w.r.t improvement, but our measure beat it by achieving a value of 0.611. However, w.r.t occlusion detection capability (Table 6 [10]), our measure gave a moderate performance of 0.172. Thus we scored the $7^{th}$ position w.r.t the other 17 measures.

Additionally, our proposed measure would not have functioned solely based on the simplified assumption that a depth map varies smoothly. Hence, our proposed measure could not have been defined based on only the entropy map of the depth map. Our experiments with the same on non-occluded regions of Wood2 yielded average 'Improvement' [10] of -0.8133, implying that it performs even worse than random chance. This may be because, in numerous locations on the depth map, like slanted surfaces or object boundaries, depth naturally varies sharply, in coherence with luminance 'L'/color. Thus, without considering 'L', our measure would have worked only on fronto-parallel surfaces.

**B. Performance evaluation of threshold detection**

We evaluate proposed threshold detection algorithm on 'all', 'nonocc' (non-occluded), and 'disc' (disparity discontinuity) regions of four rectified stereo image pairs (Table I) and their ground truth depth maps from the Middlebury Stereo Vision dataset. Table I also lists the percentage of total number of left image pixels comprising the 'all' ($P_{all}$), 'nonocc' ($P_{nonocc}$), and 'disc' ($P_{disc}$) regions.

TABLE I
CHARACTERISTICS OF THE FOUR MIDDLEBURY STEREO IMAGE PAIRS

| Stereo Pair | Resolution (pixels) | No. of Disparity Levels | $P_{all}$ | $P_{nonocc}$ | $P_{disc}$ |
|---|---|---|---|---|---|
| Tsukuba | 384 × 288 | 16 | 79.30 | 77.25 | 14.28 |
| Venus | 434 × 383 | 20 | 90.41 | 88.74 | 6.34 |
| Teddy | 450 × 375 | 60 | 97.98 | 87.50 | 24.01 |
| Cones | 450 × 375 | 60 | 96.78 | 85.29 | 27.96 |

The comparison is performed with LRC on four stereo pairs mentioned above on depth maps generated with 5×5 and 7×7 pixel blocks (same size is used for entropy calculation). All experiments were performed in Matlab on a Windows 7 PC (Intel i7-2600 quad-core 3.40 GHz CPU, 8 GB RAM). No form of task or data parallelism was used in our codes.

Performance data for both proposed thresholding method and LRC, (presented as '<proposed, LRC>' tuples) obtained from the experiments, are presented in Tables II–IV, while their execution times are presented in Table V. Our reported computation times for both methods include those consumed to compute their respective depth maps. Tables II–IV show average values of evaluation metrics by taking the size of neighborhood sizes as 5 × 5 pixels and 7 × 7 pixels.

From experimental results, we can observe the following:
1. For all stereo image pairs, the execution time is much lower for our proposed method. Since our method requires computation of depth map of only the left image, (compared to LRC, which requires computation of depth maps for both the left and right images), our method runs significantly faster than LRC, although the actual computation steps of our method are a bit more complex than LRC.
2. Both precision and accuracy are noticeably higher for our method. 'Recall' value was '100.0' for all our experiments, both in case of our method and LRC. From these, one can infer that while all of the disparity estimates which were classified as accurate by both proposed and LRC methods were actually so, (as evident from a perfect recall score for both of them), our method performed much better in identifying inaccurate estimates (due to our comparatively higher precision and accuracy, when compared to LRC).
Furthermore, many strong stereo methods [13, 14] use MRF regulation or edge-aware local cost aggregation. They may produce depth map pixels on slant surfaces having similar disparities, which are incorrect, yet satisfy our assumption. Also, LRC errors are mainly found on slant surfaces. We tested proposed and LRC method on non-occluded regions of depth maps produced by [13, 14], available on the Middlebury Stereo Evaluation page. Using neighborhood sizes of 5 × 5 and 7 × 7 for our method, average accuracy obtained was 95.43% for Teddy and Cones, and 99.83% for Tsukuba and Venus. LRC also gave accuracy above 95%, and hence, fared roughly similar to our thresholding method.

TABLE II
COMPARISON OF PROPOSED AND LRC METHODS IN 'ALL' REGIONS

| Stereo Pair | Precision (%) | Accuracy (%) |
|---|---|---|
| Tsukuba | 95.8,92.6 | 97.4,93.6 |
| Venus | 95.6,92.6 | 97.2,93.6 |
| Teddy | 88.6,85.0 | 93.3,90.4 |
| Cones | 93.4,86.3 | 96.6,91.0 |

TABLE III
COMPARISON OF PROPOSED AND LRC METHODS IN 'NONOCC' REGIONS

| Stereo Pair | Precision (%) | Accuracy (%) |
|---|---|---|
| Tsukuba | 96.3,93.0 | 97.6,94.0 |
| Venus | 96.9,93.3 | 97.9,94.1 |
| Teddy | 91.4,86.7 | 94.6,90.6 |
| Cones | 95.7,88.1 | 97.6,91.4 |

TABLE IV
COMPARISON OF PROPOSED AND LRC METHODS IN 'DISC' REGIONS

| Stereo Pair | Precision (%) | Accuracy (%) |
|---|---|---|
| Tsukuba | 92.9,87.2 | 95.7,89.4 |
| Venus | 86.1,79.2 | 91.1,84.0 |
| Teddy | 80.3,74.9 | 88.2,83.0 |
| Cones | 88.6,81.8 | 94.8,88.2 |

TABLE V
EXECUTION TIMES FOR PROPOSED AND LRC METHODS

| Stereo Pair | Proposed (seconds) (5 × 5) | LRC (seconds) (5 × 5) | Proposed (seconds) (7 × 7) | LRC (seconds) (7 × 7) |
|---|---|---|---|---|
| Tsukuba | 2.073 | 3.584 | 3.558 | 6.321 |
| Venus | 3.701 | 6.709 | 6.445 | 11.901 |
| Teddy | 9.163 | 17.539 | 15.987 | 30.908 |
| Cones | 9.137 | 17.537 | 15.957 | 31.008 |

**5. CONCLUSION AND FUTURE WORK**

We proposed a novel approach to detect stereo errors based on entropy-difference of image and depth map. Experiments show that the proposed approach is superior to most of the existing ones in many aspects. However, only w.r.t occlusion detection capability, six of the existing ones fared better than the proposed approach, on the average. Future work will improve this aspect. It will also be interesting to see whether the accuracy of proposed measure can be improved by using 'illumination' (Y) in the color space component.